\documentclass[letterpaper]{article} 
\usepackage{aaai25}  
\usepackage{times}  
\usepackage{helvet}  
\usepackage{courier}  
\usepackage[hyphens]{url}  
\usepackage{graphicx} 
\usepackage{natbib}  
\usepackage{caption} 
\usepackage{algorithm}
\usepackage{algorithmic}
\usepackage{graphicx}
\usepackage{caption}
\usepackage{tcolorbox}
\usepackage{xcolor}
\usepackage{subcaption}
\usepackage{booktabs} 
\usepackage{lineno}
\usepackage{fontawesome}

\definecolor{darkred}{RGB}{139,0,0}
\definecolor{darkgreen}{RGB}{0,100,0}
\definecolor{darkblue}{RGB}{0,0,139}
\definecolor{teal}{RGB}{0,128,128}
\definecolor{purple}{RGB}{128,0,128}
\usepackage{newfloat}
\usepackage{listings}
\DeclareCaptionStyle{ruled}{labelfont=normalfont,labelsep=colon,strut=off} 
\floatstyle{ruled}
\newfloat{listing}{tb}{lst}{}
\floatname{listing}{Listing}
\title{A Framework for Situating Innovations, Opportunities, and Challenges in Advancing Vertical Systems with Large AI Models}
\author {
    Gaurav Verma, Jiawei Zhou, Mohit Chandra, Srijan Kumar, Munmun De Choudhury
}
\affiliations {
    Georgia Institute of Technology\\
    {\{gverma, j.zhou, mchandra9, srijan, munmund\}@gatech.edu}
}

\usepackage{bibentry}

\begin{document}

\tcbset{
    colback=gray!5, 
    colframe=gray!50, 
    boxrule=0.2pt, 
    arc=4mm, 
    left=1mm, 
    right=1mm, 
    top=1mm, 
    bottom=1mm 
}



\tcbset{highlight text style/.style={enhanced,
  colframe=blue,colback=blue!10!white,arc=4pt,boxrule=1pt}}

\tcbset{highlight text style/.style={enhanced,
  colframe=red,colback=blue!10!white,arc=4pt,boxrule=1pt}}

\definecolor{vucolor}{HTML}{FFEAD8}
\definecolor{vcolor}{HTML}{C3F2D3}
\definecolor{mscolor}{HTML}{D8EDFF}
\definecolor{mcolor}{HTML}{FFD8D9}

\newtcbox{\highlightd}[1][vucolor]{
    on line,
    arc=2pt,
    colback=#1,
    colframe=#1,
    boxrule=0pt,
    boxsep=0pt,
    left=2pt,
    right=2pt,
    top=2pt,
    bottom=2pt,
}

\newtcbox{\highlightc}[1][vcolor]{
    on line,
    arc=2pt,
    colback=#1,
    colframe=#1,
    boxrule=0pt,
    boxsep=0pt,
    left=2pt,
    right=2pt,
    top=2pt,
    bottom=2pt,
}

\newtcbox{\highlightb}[1][mscolor]{
    on line,
    arc=2pt,
    colback=#1,
    colframe=#1,
    boxrule=0pt,
    boxsep=0pt,
    left=2pt,
    right=2pt,
    top=2pt,
    bottom=2pt,
}

\newtcbox{\highlighta}[1][mcolor]{
    on line,
    arc=2pt,
    colback=#1,
    colframe=#1,
    boxrule=0pt,
    boxsep=0pt,
    left=2pt,
    right=2pt,
    top=2pt,
    bottom=2pt,
}

\maketitle

\begin{abstract}
Large artificial intelligence (AI) models have garnered significant attention for their remarkable, often ``superhuman'', performance on standardized benchmarks. However, when these models are deployed in high-stakes verticals such as healthcare, education, and law, they often reveal notable limitations. For instance, they exhibit brittleness to minor variations in input data, present contextually uninformed decisions in critical settings, and undermine user trust by confidently producing or reproducing inaccuracies. These challenges in applying large models necessitate cross-disciplinary innovations to align the models' capabilities with the needs of real-world applications. We introduce a framework that addresses this gap through a layer-wise abstraction of innovations aimed at meeting users' requirements with large models. Through multiple case studies, we illustrate how researchers and practitioners across various fields can operationalize this framework. Beyond modularizing the pipeline of transforming large models into useful ``vertical systems'', we also highlight the dynamism that exists within different layers of the framework.  Finally, we discuss how our framework can guide researchers and practitioners to \textit{(i)} optimally situate their innovations (e.g., \textit{when vertical-specific insights can empower broadly impactful vertical-agnostic innovations}), \textit{(ii)} uncover overlooked opportunities (e.g., \textit{spotting recurring problems across verticals to develop practically useful foundation models instead of chasing benchmarks}), and \textit{(iii)} facilitate cross-disciplinary communication of critical challenges (e.g., \textit{enabling a shared vocabulary for AI developers, domain experts, and human-computer interaction scholars}). 
\end{abstract}

%

\section{Introduction}
Large artificial intelligence (AI) models have long served as powerful tools for advancing domain-specific research and development. Early examples include the adaptation of language embeddings (like word2vec~\cite{mikolov2013distributed}) for generating disease taxonomies \cite{ghosh2016characterizing}, as well as the use of YOLO-based image models ~\cite{redmon2016you} for conducting animal population censuses ~\cite{parham2017animal}. More recently, large models such as GPT-4o ~\cite{hurst2024gpt} and SAM ~\cite{kirillov2023segment} have demonstrated more advanced capabilities, catalyzing further innovations in domains as varied as AI tutoring ~\cite{lin2023artificial} and digital pathology ~\cite{deng2023segment}, all the way to software development ~\cite{barenkamp2020applications}. The success of these models has been propelled by the increasing scale of their architectures---rising from a few million parameters in 2013, to hundreds of millions in 2018, and now surpassing a trillion parameters in the most recent deployments~\cite{elmeleegy2024}---as well as by the sheer volume of data used for their training~\cite{bahri2024explaining}. Many widely used benchmarks, both general-purpose and domain-specific, have reached saturation or are rapidly approaching it ~\cite{chollet2024openai, phan2025humanity}. At the same time, frameworks built around these large models (such as Application Programming Interfaces; APIs) have lowered barriers to entry~\cite{schillaci2024llm}, spurring a surge in applications and attracting researchers and practitioners from a broad range of disciplines.

Despite these successes, a decade of adapting large models in diverse verticals has highlighted persistent challenges. For example, research teams have discovered that BERT and CLIP-based models can be brittle to small input variations ~\cite{verma2022robustness, ramshetty2023cross}, large language models (LLMs) often exhibit sensitivity to prompt formatting ~\cite{sclarquantifying}, and performance of large models can diminish in highly specialized settings~\cite{deng2023segment, chandra2024lived}. Moreover, AI systems sometimes struggle to effectively support diverse user groups, such as users with a lower Need For Cognition ~\cite{buccinca2021trust}. These challenges, taken together, pose bottlenecks in the vertical-adoption of the large models. The process of adapting large models to verticals demands \textit{coordinated efforts from stakeholders across varied disciplines} and continues to be an active area of research. Existing frameworks provide good starting points to address some of these problems at a more granular level---either focusing on one component of developing AI systems or a specific vertical. For instance, \citeauthor{ehsan2023charting} (\citeyear{ehsan2023charting}) present a framework to bridge the gap between social and technical aspects of developing explainable AI systems. On the other hand, focusing on a specific vertical, \citeauthor{trotsyuk2024toward} (\citeyear{trotsyuk2024toward}) present a framework to address potential misuse of AI in biomedical research. There is still an unmet need for a more \textit{comprehensive} framework comprising various components required for vertical adoption of large models and has broad \textit{applicability} towards generalizing to many verticals.

For instance, if a team were to develop an AI system to help tutor high-school children or if a different team intended to build AI that aids in the provision of psychiatric healthcare, what are the components that exist in adopting large AI models for these verticals? Which of these components would pose challenges and would need innovations from the team? On the other hand, which of these components could be addressed using existing solutions? We posit that the vertical adoption of large models can be made \textit{manageable} and \textit{modular} with a structured framework that systematically addresses the cross-disciplinary complexities. To this end, we propose a framework designed to guide both researchers and developers in optimizing large-model development and deployment by \textit{situating} their innovations, opportunities, and challenges within a clearly defined structure. 

Our proposed framework consists of 4 layers (see Figure \ref{fig:overview}), \textit{(i)} starting with \textbf{large AI models} at the bottom, \textit{(ii)} \textbf{vertical-agnostic properties}, \textit{(iii)} \textbf{vertical adaptation}, and (iv) finally, \textbf{vertical-user intermediaries}. The 4 layers represent step-wise modular abstractions involved in developing systems with large models that deliver practical value to their intended users. Drawing on case studies from multiple verticals, we discuss how to \textit{situate} innovations, challenges, and opportunities within each layer of this framework. We consider \textit{innovations} as advances in algorithms, metrics, or interface designs that resolve identified pain‑points; \textit{challenges} as specific obstacles that hinder vertical adoption of large AI models; and \textit{opportunities} as recurring unmet needs that signal room for high‑leverage solutions. We use the term \textit{situating} to indicate anchoring the contributions in one of the four layers of the framework. The overall aim of the situating contributions within the framework is to ensure that the holistic adoption of large AI models across various verticals remains effective.

Beyond describing the framework and its layers, we also highlight the inherent \textit{dynamism} among these layers (i.e., how they interact with and influence one another over time). Finally, drawing on observed trends, we present actionable recommendations that benefit researchers and developers across various verticals and disciplines. For instance, we discuss whether aspects like robustness and privacy are better addressed as vertical-agnostic properties or as a vertical-specific concerns, how scoping feedback from many verticals and intermediary problems can lead to more effective development of newer large models, when domain-specific experts can borrow modeling and interfacing techniques from others while innovating on the domain-specific data curation and evaluation methods, and how interfacing AI systems with users remains ripe with opportunities. Collectively, we believe that adopting our framework will \textit{(a)} guide optimally placed innovations, \textit{(b)} highlight potential opportunities, and \textit{(c)} enable cross-disciplinary dialogue.

\vspace{0.05in}
\noindent\textbf{\textit{Who can benefit from the framework?}} The framework is useful for interdisciplinary teams who want to adopt large AI models in their respective verticals, and for researchers who hope to position their work within a broad ecosystem for identification of cross-layer connections and translation of knowledge across contexts. For teams focused on a single vertical, the framework decomposes the adoption pipeline into modular components. Across multiple verticals, it provides a holistic view of the broader ecosystem that \textit{(a)} fosters cross-vertical exchange of innovations, opportunities, and challenges (i.e., what can teams in healthcare learn from teams in education) as well as \textit{(b)} funnels feedback from many verticals to improve the next iteration of large artificial intelligence models.

\section{A Framework for Advancing Vertical Systems With AI Models}

We developed our framework by building consensus among a group of experts with experience in creating and applying AI models across verticals such as well-being, web safety, and enterprise software. They brought expertise in artificial intelligence, human-computer interaction, social science, and healthcare, along with practical experience collaborating with leading industrial deployment teams, clinicians, non-profits, and non-governmental organizations. The group members engaged in reflective discussions, drawing on practical insights from their prior experiences deploying user-facing applications with large AI models. The identified recurring pain points were distilled into modular themes/layers that start with the underlying large models and end with users' needs. The group then discussed case studies and iteratively adapted the framework. The discussions also led to formulation of actionable recommendations for future work that aims to adopt large AI models in different verticals.

\begin{figure*}[!t]
    \centering
        \centering
        \includegraphics[width=0.65\textwidth]{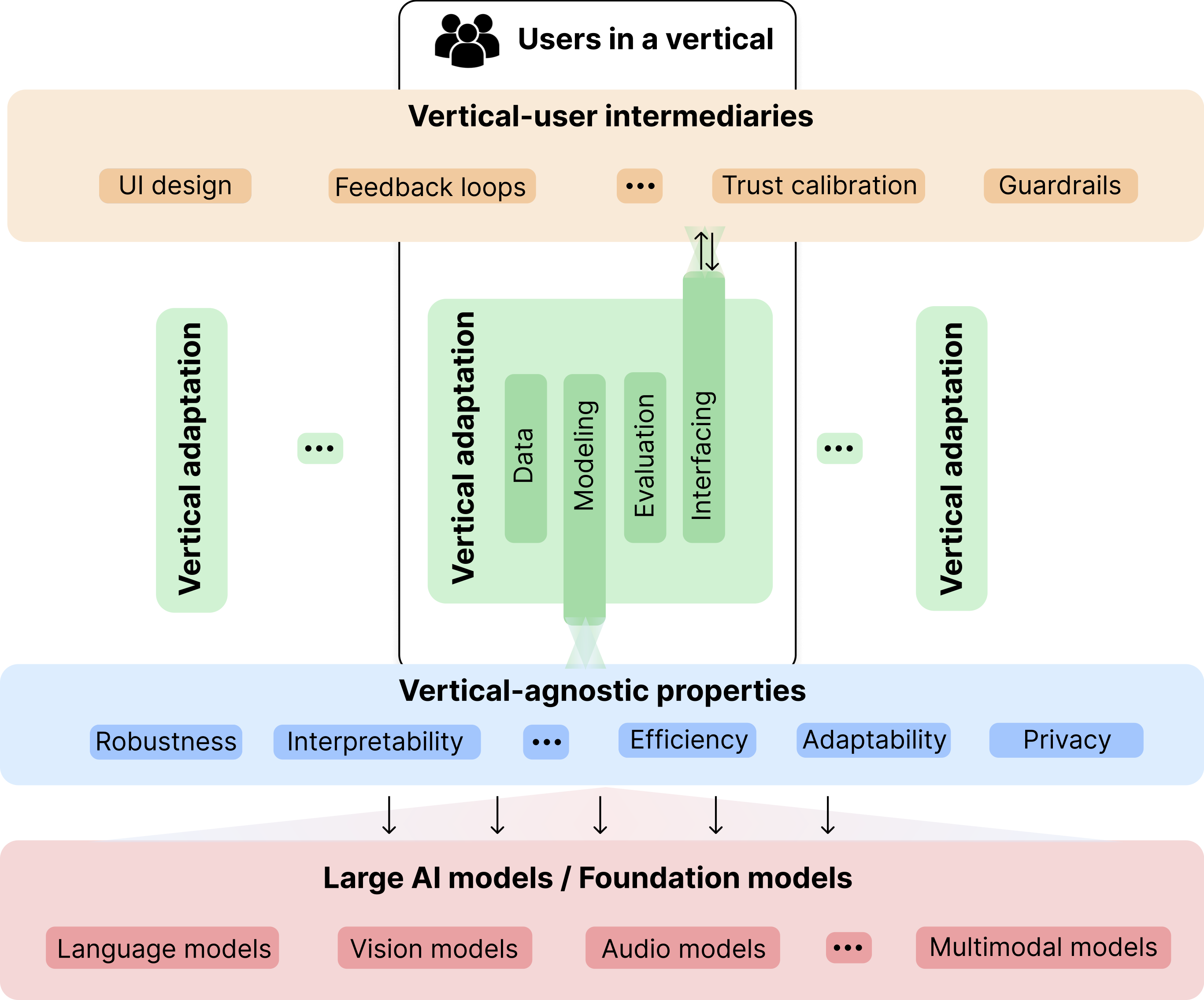}
        \caption{{Overview of the proposed framework for situating innovations, opportunities, and challenges in advancing vertical systems with large AI models; read from bottom to top. Large models form the base for vertical systems. These models need scaffolding to demonstrate properties such as robustness, interpretability, efficiency, and privacy before being useful as vertical-specific systems. Vertical-specific adaptations are required to deliver value within specific verticals. This involves curating data, designing or adapting modeling approaches, vertical-centric evaluations, and interfacing the model's outputs with users. General problems in designing interfaces and interactions between the system and users are designated as vertical-user intermediaries. The dynamism between the framework layers is noteworthy (depicted by $\downarrow$ and $\uparrow$). Over time, vertical-agnostic properties, especially those applicable to many vertical systems, could become ingrained properties in future models as development strategies evolve. Similarly, modeling for vertical adaptation could become less prominent as large models become efficiently adaptable, exemplified by the success of in-context learning with large language models. Finally, vertical-specific insights for interfacing systems with users and general interfacing techniques influence each other over time.}}
        \label{fig:overview} \vspace{-.05in}
\end{figure*}

The framework is depicted in Figure \ref{fig:overview} and described below, progressing from bottom to top, where each layer addresses specific functional aspects to achieve vertical-specific utility with large models.

\vspace{0.05in}
\noindent \textbf{1. Large AI models}: These are large-scale (in terms of training dataset size and number of parameters) AI models, more recently dubbed foundation models~\cite{bommasani2021opportunities} that power diverse applications across verticals. These include modality-specific models (e.g., language-only, vision-only) and multimodal models capable of handling multiple input modalities. Their functional utility lies in the general off-the-shelf capabilities they provide (e.g., natural language understanding, image-text alignment) and adaptability to new verticals through techniques such as fine-tuning, prompting, or in-context learning.

\vspace{0.05in}
\noindent \textbf{2. Vertical-agnostic properties}: Large models might need general scaffolding around properties like robustness, interpretability, efficiency, and privacy, before they are useful as vertical-specific systems---such problems are considered vertical-agnostic properties. While improvements along these aspects generally benefit many verticals, nonetheless, some of these aspects may \textit{also} require vertical-specific considerations.

\vspace{0.05in}
\noindent \textbf{3. Vertical adaptation}: Designed for delivering specific value within verticals such as healthcare, web safety, and education, vertical-specific adaptations involve integrating large models' capabilities with vertical-specific data, modeling, evaluation, and interfacing the capabilities with end-users of the vertical.

\vspace{0.05in}
\noindent \textbf{4. Vertical-user intermediaries}: These address vertical-agnostic challenges in interfacing systems built with large models with the end-users, focusing on aspects like trust calibration, feedback loops, and dynamic interfaces. While some interfacing challenges could be vertical-specific, others are broadly applicable challenges.

The following section presents case-studies that apply this framework to two verticals --- healthcare and education.

\section{Applying the Framework: Case Studies}

As we describe the innovations, opportunities, and challenges that exist in the vertical adoption of large models and situate them in our framework, two questions guide our efforts: ``Who are we trying to help?'' and ``What do they care about?'' These considerations gain prominence at higher levels---where user impact is tangible---but inform decisions throughout every layer of the framework. Table \ref{tab:case-studies} depicts some of the example questions within different layers of the framework. These example problems show that each vertical has different data, modeling, and evaluation needs, yet they share cross-cutting challenges that exist between large AI models and vertical systems as well as between vertical systems and users.
 
\vspace{0.05in}
\noindent\textbf{Large AI models}: The architectural scale and the pre-training dataset size have resulted in remarkable off-the-shelf capabilities of the large AI models, measured by their success on continually evolving benchmarks\cite{chollet2024openai, liangholistic}. While there has been a push towards even larger models trained on huge datasets, equipping them with the ability to support inputs in multiple modalities --- language, vision, audio, and in some cases even sensor data~\cite{moon2024anymal}, has the potential to unlock new applications across many verticals. For instance, multimodal models could, in principle, process radiology scans along with diagnostic questions~\cite{bhayana2024chatbots}, raw electrocardiogram (ECG) signals for health data analytics~\cite{quer2024potential}, provide voice-based tutoring~\cite{katsarou2023systematic}, and write and review lengthy codebases~\cite{bairi2024codeplan}. While the development of these AI models offers an affordance for multimodal input, questions remain around how \textit{well the multimodal LLMs can reason over the non-textual forms of data}\cite{tong2024eyes, verma2024cross}, which requires further work within this layer of the framework. Advances that ensure multimodal LLMs indeed model all modalities reliably will ensure greater off-the-shelf capabilities in the future iterations of these models.

\vspace{0.05in}
\noindent\textbf{Vertical-agnostic properties}: As we consider applying large AI models, the first set of problems are vertical-agnostic properties that apply to most verticals and pave the path for effective consideration of vertical-specific aspects. For instance, are multimodal LLMs robust to the plethora of plausible and realistic variations in user-provided inputs, given that it is unreasonable to assume users will constrain their inputs to the margins of the training distribution~\cite{ramshetty2023cross, verma2022robustness, nookala2023adversarial}? Will these models handle the personally identifiable information (PII) already encoded from its pre-training corpus~\cite{carlini2021extracting} and the sensitive data provided by users while using it for iterative refinement (e.g., from patients~\cite{pan2024adaptive} or students~\cite{yang2024ensuring})? Relatedly, can large AI models provide interpretable predictions that foster transparency~\cite{stiglic2020interpretability}? Addressing these problems as foundational steps will overcome bottlenecks across many verticals and enable a more effective use of the remedial techniques. Additionally, addressing these problems in the proximity of the large AI models layer could lead to integration of the remedial techniques in the development of upcoming large AI models; we elaborate on this in Section \ref{sec:dynamism}. We now move on to situating vertical-specific innovations, opportunities, and challenges within the framework, starting with healthcare and then exploring education. We chose to focus on healthcare and education verticals because they represent high-stakes domains with distinct yet complementary challenges: both verticals focus on building user-centered systems with healthcare demanding precision and rigorous safety measures while education prioritizing engagement and personalization to cater to diverse pedagogical needs of the learner.

\begin{table*}[!t]
\centering
\renewcommand{\arraystretch}{1.2}
\begin{tabular}{@{}c@{}}
\toprule
\parbox{0.95\linewidth}{
    \begin{center}
        \highlightd{\textbf{Vertical-User Intermediaries}}\\[4pt]
    \end{center}
    
    {\highlightd{Trust calibration}: How can AI systems effectively communicate their capabilities and limitations to users to prevent issues of overreliance or unwarranted skepticism?}\vspace{0.05in}\\
    {\highlightd{Feedback loops}:How can real-time user feedback be systematically captured and integrated into iterative refinement processes for large AI models?}\vspace{0.05in}\\
    {\highlightd{Dynamic interfaces}: How can interface designs dynamically accommodate users with varying cognitive engagement preferences, optimizing usability for both high and low Need For Cognition (NFC) individuals?}\\
}\\
\midrule
\parbox{0.95\linewidth}{
    \centering
    \begin{tabular}{@{}p{8.2cm}|p{8.3cm}@{}}
        \toprule
        \begin{center}
            \textbf{\highlightc{Vertical Adaptation} in \underline{Healthcare}} 
        \end{center}
            &
        \begin{center}
            \textbf{\highlightc{Vertical Adaptation} in \underline{Education}}
        \end{center}\\
        
        \highlightc{Data}: {What methodologies can be used to curate specialized datasets of medical dialogues necessary for accurately tuning multimodal models for clinical use?}\vspace{0.05in}             & \highlightc{Data}: {How should educational data be curated to reflect diverse pedagogical strategies and learner demographics to improve AI-assisted tutoring?} \vspace{0.05in}       \\
        \highlightc{Modeling}: {How can AI models effectively incorporate temporal patient history data to improve accuracy in clinical diagnostics?} \vspace{0.05in}            & \highlightc{Modeling}: {What modeling techniques can efficiently adapt AI systems to individual learner needs and preferred pedagogical strategies?} \vspace{0.05in}   \\
        \highlightc{Evaluation}: {Which clinical standards and metrics most effectively measure the quality and reliability of AI-generated outputs in healthcare settings?} \vspace{0.05in}         & \highlightc{Evaluation}: {How can AI system evaluations accurately assess their impact on learner motivation, engagement, and educational outcomes beyond correctness of content?} \vspace{0.05in} \\
        \highlightc{Interfacing}: {How can AI systems best support collaboration with clinicians?}   & \highlightc{Interfacing}: {How can AI tutoring systems effectively ground responses within the immediate learning context and visual perspective of the student?} \\
        \bottomrule
    \end{tabular}
}\\
\midrule
\parbox{0.95\linewidth}{
    \begin{center}
        \textbf{\highlightb{Vertical-Agnostic Properties}}\\[4pt]    
    \end{center}
    
    {\highlightb{Robustness}: How robust are AI models to realistic \& plausible variations in user inputs spanning multiple modalities?} \vspace{0.05in}\\
    {\highlightb{Privacy}: How effectively can AI models ensure the privacy \& security of personal and sensitive data?} \vspace{0.05in}\\
    {\highlightb{Interpretability}:  In what ways can AI models be made to provide interpretable predictions to improve transparency?}\\
}\\
\midrule\midrule\\
\textbf{\highlighta{Multimodal Large Language Models}}\vspace{3mm}\\
{How can modalities beyond language (visual, audio, sensor data) be reliably processed with large AI models?}\\
\bottomrule
\end{tabular}
\caption{{Operationalizing the framework for adoption of large AI models in the healthcare and education verticals; read from bottom to top. Each question is an example of existing/potential innovations, challenges, and opportunities in vertical adoption of large artificial intelligence models (in this case, multimodal large language models).}}
\label{tab:case-studies}
\end{table*}

\vspace{0.05in}
\noindent\textbf{Vertical adaptation: (a) MLLMs for healthcare}: One of the well-explored clinical applications of multimodal deep learning models, including the recent multimodal LLMs, is as an assistant for radiologists. The models can help clinicians make diagnoses via conversational-assistance as well as writing medical reports~\cite{johri2025evaluation, zhang2024generalist}. Even though off-the-shelf LLMs encode medical knowledge~\cite{singhal2023large}, they need to be vertically adapted to acquire diagnostically useful information from natural conversations with primary care providers. This vertical adaptation involves \textit{curating the right data of diagnostic conversations}, which would require close involvement of domain-experts~\cite{tu2024towards}. Beyond curating the right data, off-the-shelf LLMs do not demonstrate properties that are required to accurately model the data--for instance, history-taking~\cite{tanno2024collaboration, tu2024towards}. In the context of radiology, this would involve \textit{equipping the underlying multimodal LLMs with temporal modeling capabilities}~\cite{bannur2023learning}. To evaluate whether the modeling approach is effective on the curated data, vertical-agnostic evaluations may not suffice. For instance, Yu et al. (2023)~\cite{yu2023evaluating} demonstrate that generic `natural language generation' metrics are not effective in capturing clinically pertinent differences between AI-generated radiology reports and those written by experts, and \textit{propose meaningful metrics to guide future research} in this vertical. The evaluations uncovered that while both experts and AI systems can make mistakes while generating radiology reports alone, the instances of inaccuracies decrease when experts and AI work in collaboration to fix the errors~\cite{tanno2024collaboration}. Nonetheless, a crucial \textit{challenge remains unaddressed when interfacing the AI systems with clinicians} -- the collaboration loses effectiveness when the expert either overly relies on the AI predictions~\cite{seah2021effect, rajpurkar2020chexaid} or is excessively critical of them~\cite{agarwal2023combining}.

\vspace{0.05in}
\noindent\textbf{Vertical adaptation: (b) MLLMs for Education}: Large AI models, including multimodal LLMs, are already being used by students across the globe to assist with their learning~\cite{zhu2024embracing}, demonstrating their promise in tutoring. Even though their ad hoc capabilities are noteworthy, Macina et al. (2023) ~\cite{macina2023opportunities} note that their systematic impact on ``tutoring has largely remained unaffected''. A central challenge in this vertical concerns \textit{curating data that captures diverse pedagogical strategies}, covering a broad range of topics, learner demographics, and instructional modes encountered in real classrooms~\cite{jurenka2024towards}. From a modeling perspective, besides fine-tuning (which may prove to be an inefficient strategy for adapting to different definitions of what constitutes effective pedagogy), it is \textit{advantageous to allow learners to specify desired attributes across pedagogical dimensions and have the model reflect them}~\cite{team2024learnlm}. The \textit{evaluation of such models should be grounded in learning science}, which often prioritizes motivating and promoting engagement from the learner, and not just ``giving the right answer''\cite{foster2023edtech}. It is also crucial that the \textit{interfaces support learner-tutor interactions such that the conversations are grounded in what the student ``sees''}, which is one of the core principles of student-centric pedagogy.

\vspace{0.05in}
\noindent\textbf{Vertical-user intermediaries}: Even though there are vertical-specific considerations involved in interfacing systems with users, some of these challenges are frequently encountered across many verticals. For instance, as we noted, clinician-AI collaboration tends to become less useful when the capabilities and limitations are not adequately understood by the experts. Similar observations around `algorithm aversion or appreciation' exist across  other verticals~\cite{dietvorst2015algorithm, logg2019algorithm, qian2024take}. Given the broad nature of the underlying challenge, it is effective to work on \textit{trust calibration} as a vertical-agnostic vertical-user intermediary.  E.g., introducing uncertainty expressions (``I'm not sure, but...'') in AI-generated responses leads to decreased over-reliance and calibrated trust among users~\cite{kim2024m} -- an insight that can benefit many verticals, including generative information retrieval, healthcare assistance, and educational tutoring. Similarly, designing interfaces that capture real-time, in-situ, and implicit user feedback, will unlock iterative refinement of underlying systems across many verticals\cite{shi2024wildfeedback}. Furthermore, interfaces that dynamically adapt to deliver the optimal experience to a group of users with diverse Need For Cognition levels (NFC; a personality trait that considers users' tendency to engage with cognitive activities\cite{cacioppo1982need}) is an opportunity that will benefit many verticals. More specifically, for decades, researchers have observed that users with higher NFC levels benefit significantly from complex interfaces, while others struggle to adopt them\cite{carenini2001analysis} --- a pattern that spans many verticals and has continued with recent generative artificial intelligence technologies~\cite{toner2024artificial,buccinca2021trust}.

\section{Dynamism Between Layers of the Framework}
\label{sec:dynamism}
The framework is structured in layers to support \textit{modularity} in addressing innovations, opportunities, and challenges. Yet these layers do not serve the purpose of \textit{rigidly demarcating} the underlying problems. Instead, as we discuss in this section, they exhibit extensive cross-layer interactions and mutual influences---referred to here as ``dynamism''. This section provides concrete examples that (a) illustrate this dynamism and (b) show how the dynamism can foster distributed-yet-collaborative synergy in AI's development.

Let us start with \textbf{vertical-specific considerations influencing vertical-agnostic properties}. As practitioners and researchers explore the applications of large models in different verticals, they highlight the successes as well as shortcomings. Certain aspects such as lack of robustness, poor handling of private data, and lack of efficient adaptability are frequently \textit{encountered/identified} across verticals and make their way to being more effectively \textit{addressed} in a vertical-agnostic manner. When addressing these challenges in a vertical-agnostic manner, \textbf{vertical-agnostic solutions can influence the development of the next iterations of large AI models}. Matryoshka representations~\cite{kusupati2022matryoshka} present a strong case study for this. In many real-world search systems (e.g., patient records, legal reviews, and educational video search), relevant items are often retrieved by computing similarity between neural representations. At large scales---hundreds of millions of items---this must be efficient. Kusupati et al. (2022)~\cite{kusupati2022matryoshka} introduced ``Matryoshka doll'' representations, where $14\times$ smaller embeddings match the performance of full-size ones for classification and retrieval. This innovation supports flexible representation sizes, enabling efficient retrieval and classification in a vertical-agnostic manner, even though the initial inadequacies were identified during many vertical-specific adaptations. Although originally the effectiveness of Matryoshka representations was demonstrated on ImageNet-scale datasets, it was later adopted by OpenAI and others to train large AI models (e.g., \texttt{\small{text-embedding-3}}) on web-scale data~\cite{OpenAI2024}.

Akin to the dynamism between vertical-specific considerations, vertical-agnostic properties, and large AI models, certain challenges may also \textbf{shift from interfacing-related aspects within verticals to broader vertical-user intermediaries}. Consider AI hallucinations in high-stakes fields like healthcare, where unreliable responses can lead to adverse outcomes and erode trust, prompting research on cognitive forcing interventions in medical AI-assisted decision-making~\cite{buccinca2021trust}. With LLMs now prevalent in many domains (e.g., web retrieval, law, education, and mental healthcare), hallucination has become a key challenge. This has spurred work on vertical-user intermediaries, including quantifiable or language-based uncertainty cues~\cite{xiongcan, kim2024m} and referencing source documents for verifiable claims~\cite{gao2023enabling}, with early evidence of reduced user over-reliance on AI outputs~\cite{bo2024rely}.

The \textbf{dynamism between layers} is inherently desirable as it \textbf{facilitates the distributed-yet-collaborative development of AI systems}. The large AI models are often developed in highly resource-rich environments, while the practitioners and researchers who develop vertical-specific insights that trickle down to the bottom layers of the framework are often situated outside of these selective environments. Here, two points are worth noting: first, to deliver benefits within a vertical using large AI models, it is critical to actively engage with domain experts who have curated specialized data, engineered novel methods, or designed meaningful evaluations for those verticals. This enables specialized insights that arose from vertical-specific explorations, while simultaneously embedding them in more general-purpose model architectures. Second, it is \textit{incorrect} to assume that vertical-specific explorations merely \textit{piggyback} on advances in large AI models. Rather, vertical-specific advances play a critical role in the development of large AI models. When used carefully, these vertical-specific insights can make general-purpose models more capable and speed up the release of solutions for specific user needs.

\section{Intended Framework Outcomes}
Developing on the trends that illustrate the dynamism between layers of the framework and its collaborative-yet-distributed nature, we discuss how the framework can aid along the following axes: encouraging researchers to optimally situate or borrow innovations, discovering overlooked opportunities, and engaging in a structured cross-disciplinary dialogue. The actionable recommendations were derived through an iterative process, where the authors reflected on recurring pain points encountered across different applications of large AI models and deliberated on steps that could enable their effective integration.

\subsection{Optimally situating and borrowing innovations}

One of the key intended outcomes of the framework is to help situate the innovations such that they have increased potential for impact and also promote optimal use of resources. An important aspect of this is to \textbf{consider when vertical-specific innovations, particularly on the modeling and interfacing front, could have broader impact across many verticals}. For instance, studies across many verticals have found that adapting LLMs on benign datasets for healthcare, education, and law compromises their safety --- making them more likely to respond to potentially harmful prompts~\cite{qifine}. Individual research teams across verticals therefore go through additional steps to first \textit{quantify the extent of compromise in safety upon fine-tuning} and then, if the extent is unacceptable, \textit{improve the compromised safety of vertically-adapted LLMs}~\cite{niknazar2024building, de2024chatbots}. However, as opposed to investigating and addressing the underlying issues independently in many verticals, Peng et al. (2024) study the loss of safety in adapted LLMs---across several LLMs and datasets---to propose a metric for safety in LLM adaptation by visualizing its safety landscape~\cite{peng2024navigating}. The metric could be used across verticals to determine whether the LLM adaptation has compromised safety and the extent to which remedial strategies are required. This illustrates how addressing challenges that are encountered across several verticals as a vertical-agnostic properties could be more optimal in terms of impact and resource allocation. 

Similar examples exist in how vertical-specific interfacing challenges could point to broadly applicable problems that are optimally addressed as vertical-user intermediaries. As a specific example, the tendency of LLM-based chatbots to neglect crucial aspects of user interactions as they primarily focus on outcomes has been noted in many verticals, including their applications in therapeutic conversations~\cite{zhou2024s}, information seeking~\cite{sharma2024generative}, writing code~\cite{bajpai2024let}, or humanitarian frontline negotiations~\cite{ma2024chatgpt}. While vertical-specific strategies have been proposed (like inducing Gricean maxims to structure code-related interactions~\cite{bajpai2024let}), it is promising to assess whether such strategies generalize across verticals as a vertical-user intermediary. If such a challenge can be addressed in a vertical-agnostic manner, it could enable more effective vertical-user interactions across many verticals. It is worth emphasizing that situating innovations as vertical-agnostic properties (whether in modeling or interfacing) does not aim to discourage vertical-specific innovation as it is the latter that provides insights into what could work (snowballing effect). The objective here is to encourage researchers and practitioners to consider the potential impact of their innovation and aid others in adopting it to establish vertical-agnostic generalization. 

Conversely, optimality also requires that researchers and practitioners \textbf{adopt vertical-agnostic strategies when they are effective in their specific verticals}. For instance, while many vertical-specific works note that prompt engineering is often ad hoc, solutions like DSPy~\cite{khattab2023dspy} provide a structured, vertical-agnostic approach to optimizing prompts, thereby addressing common challenges across multiple domains and saving resources.

\subsection{Uncovering overlooked opportunities}
Guided by the philosophy of ``turning frequent failures into signals'', the framework can also help uncover overlooked opportunities. A compelling example exists in the development and adoption of large language models for languages in South Asia. Individual teams have noted the limitations of LLMs trained on predominantly English data when they are used in non-English languages for applications in different verticals~\cite{jin2024better, kumar2024socio, verma2022overcoming}. Such LLMs also lack the socio-cultural awareness to adapt to the needs of users in South Asian regions~\cite{pawar2024survey}. These failures have highlighted an opportunity for collaborative and participatory research to develop new LLMs for South Asian languages, such as BharatGPT~\cite{bharatgpt2025} and SeaLLMs~\cite{nguyen2023seallms}. Since their deployment, these models have been adopted across many verticals by businesses, governments, and non-profits~\cite{corover2025}.  By identifying the pattern among the failures in vertical adoption and addressing the major issues at the large AI model-layer of the framework, the vertical-agnostic and vertical-specific layers can build on top of advanced off-the-shelf capabilities of the newer models. Such \textbf{coordinated and targeted efforts also ensure that the burden of developing large AI models that are effective in specific verticals does not fall on individual teams} as the process is restrictively resource intensive and carries environmental costs~\cite{strubell2020energy}. 

Another bottleneck that exists in the vertical adoption of large models is careful handling of sensitive data -- including health data~\cite{pan2024adaptive}, student data~\cite{yang2024ensuring}, and proprietary workflows~\cite{tangprivacy}. Current large models readily allow adaptation via in-context learning or custom system instructions, which pose the risk of exposing sensitive data via jailbreak attacks~\cite{liu2023jailbreaking}. However, Rajendran et al. (2024)~\cite{rajendran2024learning} argue that cross-cohort cross-category integration --- ``\textit{the process of combining information from diverse datasets distributed across distinct, secure sites}'' --- is important for adopting large models in verticals like healthcare. This provides an \textbf{opportunity to develop large AI models or vertical-agnostic properties that facilitate adaptability while securely handling data without compromising on key performance metrics}. Training large models for adaptability via approaches like meta-learning~\cite{verma2024adaptagent} or providing \textit{secure} adaptability as a service~\cite{tangprivacy} are some opportunities to accelerate vertical adoption of large AI models. 

Beyond modeling-related opportunities, there are significant \textbf{opportunities on the human-AI interaction and interface design fronts, both in vertical-specific as well vertical-agnostic settings}. The large-scale user-adoption of AI is still relatively nascent~\cite{denhouter2025, abril2024} and while modeling-related problems (both vertical-specific and vertical-agnostic) are being actively addressed by newer iterations of the large models, the interfacing of these capabilities with the intended users is still a major challenge. For instance, beyond chatbot-like interfaces, recent studies show structured media such as notebooks provide a flexible interface for incrementally creating and consuming information, which is also effective for clinically mandated documentation standards~\cite{cheng2024biscuit, wangStickyLand2022, adler2022meeting}. Additionally, micro-prompting and using interactive graphical objects (an interaction paradigm that can be applied to many verticals) has shown to enhance user satisfaction in interactions between human and AI systems~\cite{suh2023sensecape, jiang2023graphologue,butler2024nfw}. 

\subsection{Communicating cross-disciplinary challenges}

Achieving real-world impact with large AI models demands coordinated expertise spanning AI architectures, high-performance hardware systems, data curation, domain knowledge (including regulatory considerations), and human-computer interaction. \textbf{Our framework offers a shared language for communicating these varied technical and societal needs}. It helps each group pinpoint bottlenecks and future directions across the layers. For instance, researchers who develop a modeling or interface solution in one vertical can highlight its potential generalizability, bringing it to the attention of others working on similar challenges in different verticals. Relatedly, researchers who develop vertical-agnostic solutions that address challenges across many verticals should persuade their adoption across verticals. If these solutions prove effective repeatedly across many verticals, large-model developers should incorporate them into next-generation models and architectures so they become standard off-the-shelf capabilities.

By helping researchers situate their efforts within a common structure, the framework promotes streamlined collaboration among AI developers, vertical experts, and human-computer interaction researchers, ensuring that large models evolve into useful and trustworthy tools across a wide range of real-world applications. The framework also helps {avoid the pitfalls of viewing large AI models as self-sufficient solutions}~\cite{blodgett2021risks, bommasani2021opportunities}. At the same time, it \textbf{encourages vertical-specific teams to articulate their domain's unique requirements and allows large-model developers to spot recurring problems that merit general-purpose fixes}. 

\section{Discussion}

\noindent\textbf{Positioning with respect to other frameworks}: As mentioned earlier, most existing frameworks focus on specific facets of adopting large AI models into verticals. For instance, ~\citeauthor{ehsan2023charting} (\citeyear{ehsan2023charting}) chart the sociotechnical gap in explainable artificial intelligence, presenting a framework that conceptualizes the gap between model outputs and human interpretability needs. Similarly, ~\citeauthor{goldstein2024ppou} (\citeyear{goldstein2024ppou}) present a framework for estimating the malicious use of systems built with advanced AI models. In the domain of conversational search, \citeauthor{azzopardi2024conceptual} (\citeyear{azzopardi2024conceptual}) develop a conceptual model of agent–user interaction, mapping out the dialogue acts and decision points that drive information-seeking conversations. Moving beyond facets like explainability and interfacing, at a systems level, \citeauthor{AppliedLLMs2024} (\citeyear{AppliedLLMs2024}) provide a list of engineering-focused best-practices for LLM-based applications, emphasizing components like rigorous evaluation, retrieval-augmented generation, fine-tuning, and guardrails to integrate large models into products. In the same vein, institutions have also put forth adoption guides: a U.S. Department of Energy report (~\citeyear{DOE_AITO_2022}) underscores the need for robust data pipelines, high-performance computing, and reproducible model infrastructures to accelerate AI uptake in critical domains. Likewise, the NIST AI Risk Management Framework (~\citeyear{NIST_AIRMF_2023}) focuses on processes for identifying and mitigating AI system risks, defining functions such as Govern, Map, Measure, and Manage to ensure trustworthy development and deployment. Industry frameworks like IBM’s ``AI Ladder'' (~\citeyear{stryker2025ai}) similarly prescribe sequential steps (e.g., modernize, collect, organize, analyze, infuse) to guide organizations in scaling AI from data preparation to integration into workflows.

Each of these works offers valuable conceptual frameworks, but notably each isolates a specific sub-problem or component of AI adoption into verticals --- be it explainability, user interfacing, engineering best practices, infrastructure needs, or product management cycles --- rather than providing a unified framework. In contrast, our layered framework offers a holistic and modular structure that ties these aspects together. It spans from the core large-model layer, through vertical-agnostic properties and vertical-specific adaptation, up to user-facing intermediaries, explicitly situating innovations and challenges in context. Within this integrated view, we present several perspectives that could enable an ecosystem of efficient and effective vertical adoption of AI models --- e.g., how improvements in one layer (for example, a robustness technique at the model layer or a novel interfacing paradigm at the user-intermediary layer) can propagate benefits across other layers and across domains. Cross-cutting concerns, such as data privacy, fairness, or trust, are better addressed at right layers such that they are not repeatedly ``solved'' in silos.

By providing a shared vocabulary and clear abstraction boundaries, the framework enables researchers and practitioners to communicate and build upon each other's advances. Our framework not only advocates to reduce redundant effort (teams in different verticals can reuse solutions or insights from analogous layers) but also establishes a basis for diving deeper into aspect-specific frameworks as needed.  For instance, an explainability framework (as in the framework by \citeauthor{ehsan2023charting}) slots naturally into our vertical-user intermediary layer, and a conversational interaction model (as in \citeauthor{azzopardi2024conceptual}'s framework) can be viewed as a vertical-specific interface component.  In essence, the layered approach ``situates'' specialized innovations within a bigger picture. All of these qualities address the unmet need for an integrated conceptual models for AI adoption: one that modularly encompasses the full pipeline of transforming a large AI model into a domain-specific, user-facing system, and thereby complements prior frameworks.

\vspace{0.05in}
\noindent\textbf{Adaptability with the changing interaction paradigm}: While our framework primarily focuses on human-AI interactions, it is inherently adaptable to emerging ways of interactions with large models. For instance, Model Context Protocol (MCP) is an open standard recently designed to facilitate interactions between large models and external tools~\cite{anthropicIntroducingModel}. As agentic-AI workflows gain prominence, understanding signals and feedback from model-tool interactions becomes crucial for further improvements. However, such signals could differ from traditional measures commonly used in human-AI interaction paradigm (such as user trust) and may include newer measures such as agentic-coordination capabilities, and tool-usage efficiency. Our framework provides the agility to incorporate such changes. For example, in this case, an additional layer parallel to the vertical-user intermediary layer could be introduced which specifically adapted to analyze multi-agent system interactions.

\vspace{0.05in}
\noindent\textbf{Limitations and future directions}: Our framework, while providing a valuable starting point for situating innovations, opportunities and challenges in vertical adoption of large AI models, has multiple limitations. The framework centers on end-users and their direct interactions with AI systems, leaving out broader social, institutional, and policy contexts. In the context of healthcare and well-being, De Choudhury et al. (2023)~\cite{de2023benefits} adopt a Social Ecological Model to explore how AI can influence not just individuals but also caregiving institutions and society at large. While these wider sociotechnical considerations are crucial for fully understanding AI's impact --- especially in sensitive areas like mental health and telehealth --- they fall beyond the scope of our current user-focused framework. Future work could integrate our framework with broader ecological models to capture the dynamism across multiple stakeholders and social layers, thereby offering a more holistic view of how AI systems both shape and are shaped by larger sociotechnical contexts. Furthermore, since our framework aims to cover the development pipeline that transforms large AI models into vertical systems, it does not dive deep into individual challenges. For instance, prior studies have developed frameworks around specific aspects like robustness, interpretability, explainability, and human-AI interaction and communication~\cite{li2023trustworthy, fragiadakis2024evaluating, bansal2024challenges}. Future work could involve a large-scale study that balances high-level coverage of the entire pipeline with deeper dives into each sub-issue. This would require a multi-disciplinary collaboration of experts in those sub-issues, ensuring that the final systems remain grounded in real user needs while thoroughly addressing the nuanced challenges in each aspect of deployment.

\section{Conclusion}
Our framework addresses the crucial gap between the remarkable capabilities of large AI models and the complex, context-specific needs that arise in real-world verticals. By modularizing the development pipeline into four layers --- large AI models, vertical-agnostic properties, vertical adaptation, and vertical-user intermediaries, our framework offers a structured way to identify, situate, and address the challenges encountered when building practical AI systems. We also highlighted the dynamic interplay among these layers: insights from domain-specific challenges loop back into model-level improvements, and widespread interface challenges become generalizable solutions for human-AI interactions. To that end, the framework provides actionable guidance for effectively placing innovations, spotting opportunities that might otherwise be overlooked, and coordinating across disciplines to precipitate vertical-specific utility.

Large AI models hold tremendous promise but are not plug-and-play solutions that immediately translate into user-facing impact. Our work underscores the healthy dynamism needed to meaningfully apply these models: it calls on AI developers to incorporate feedback from vertical deployments (instead of merely chasing benchmarks), and it urges vertical-focused researchers to recognize where their innovations could broaden into vertical-agnostic improvements. Ultimately, this coordinated and interdisciplinary effort will ensure that large AI models truly deliver on users' needs.

\section{Acknowledgments}
This research is based upon work supported in part by NSF grants CNS-2154118, ITE2137724, ITE-2230692, CNS2239879, NIH grants R01 MH117172 and P50 MH115838, Defense Advanced Research Projects Agency (DARPA) under Agreement No. HR00112290102 (subcontract No. PO70745), CDC, and funding from Microsoft and the American Foundation for Suicide Prevention (AFSP). Any opinions, findings, and conclusions or recommendations expressed in this material are those of the author(s) and do not necessarily reflect the position or policy of NSF, NIH, DARPA, DoD, SRI International, CDC, Microsoft, or AFSP, and no official endorsement should be inferred. Gaurav is partly supported by the JPMorgan AI Research PhD Fellowship and the Snap Research Fellowship.


\bibliography{aaai25}


\end{document}